\documentclass[conference]{IEEEtran}
\IEEEoverridecommandlockouts

\usepackage{cite}
\usepackage{amsmath,amssymb,amsfonts}
\usepackage{algorithmic}
\usepackage{graphicx}
\usepackage{textcomp}
\usepackage{xcolor}
\usepackage{hyperref}
\usepackage{color}
\usepackage{tabularx}
\usepackage{booktabs}
\usepackage{multirow}
\usepackage{url}
\usepackage{float}
\usepackage{makecell}
\usepackage[labelformat=simple]{subcaption}

\usepackage[mathscr]{eucal}

\def\euca#1{\boldsymbol{\mathscr #1}}


\usepackage{xpatch}
\xpatchcmd\IEEEkeywords{---}{-}{}{}

\makeatletter
\renewcommand{\fnum@figure}{Figure~\thefigure}
\makeatother

\usepackage{caption}
\def\BibTeX{{\rm B\kern-.05em{\sc i\kern-.025em b}\kern-.08em
    T\kern-.1667em\lower.7ex\hbox{E}\kern-.125emX}}

\begin{document}

\title{{\bfseries\Large Quantized Rank Reduction: A Communications-Efficient Federated Learning Scheme for Network-Critical Applications}\\

\author{Dimitrios Kritsiolis \IEEEauthorblockN{and Constantine Kotropoulos}
\IEEEauthorblockA{\textit{Department of Informatics} \\
\textit{Aristotle University of Thessaloniki}\\
Thessaloniki 54124, Greece\\
email: \{dkritsi, costas\}@csd.auth.gr}
}}

\maketitle

\begin{abstract}
Federated learning is a machine learning approach that enables multiple devices (i.e., agents) to train a shared model cooperatively without exchanging raw data. This technique keeps data localized on user devices, ensuring privacy and security, while each agent trains the model on their own data and only shares model updates. The communication overhead is a significant challenge due to the frequent exchange of model updates between the agents and the central server. In this paper, we propose a communication-efficient federated learning scheme that utilizes low-rank approximation of neural network gradients and quantization to significantly reduce the network load of the decentralized learning process with minimal impact on the model's accuracy.
\end{abstract}

\begin{IEEEkeywords}
\textit{federated learning; Tucker decomposition; SVD; quantization.}
\end{IEEEkeywords}

\section{INTRODUCTION}

As artificial intelligence and machine learning evolve, new computational paradigms are emerging to address the increasing demand for privacy, efficiency, and scalability. One such approach is Federated Learning (FL), a decentralized learning technique that enables model training across multiple devices or agents without requiring direct data sharing \cite{konevcny2016federated} \cite{zhang2021survey}. In FL, end devices train their model using local data and send model updates to the server for aggregation rather than sharing raw data. This approach enhances data privacy while allowing the server to refine the global model based on updates from multiple devices. FL is a key enabler of artificial intelligence in mobile devices and the Internet of Things (IoT) \cite{lim2020federated}.

One of the key challenges in FL is the significant communications overhead, which does not scale efficiently as the number of participating devices increases \cite{asad2023limitations}. The just-described issue becomes even more pronounced in deep learning, where models consist of voluminous parameters that must be shared by each client with the server at every training iteration. As a result, the communication bottleneck diminishes the advantage of distributed optimization, slowing the overall training process and reducing the efficiency gains expected from decentralized learning \cite{kairouz2021advances} \cite{li2014communication}. To address this issue, we aim to compress and quantize the updates sent by clients, thereby mitigating the effects of communication overhead without significantly deteriorating the model's performance.

Before explaining the compression and quantization techniques, we formally introduce the distributed learning problem \cite{arjevani2015communication} solved by FL, i.e., 
\begin{equation}
    \min_{\pmb{\theta}} f(\pmb{\theta})=\min_{\pmb{\theta}} \sum_{c \in \mathfrak{C}} f_c(\pmb{\theta}) \;\;\; \text{with} \;\;\; f_c(\pmb{\theta}) := \sum_{n=1}^{N_c} J(\pmb{X}_{c, n}; \pmb{\theta}),
    \label{eq1}
\end{equation}
where $\pmb{\theta}$ denotes the parameters of the central model being trained, $\mathfrak{C}$ is the set of clients participating in FL with $|\mathfrak{C}| = C$, $\pmb{X}_{c, n}$ is the $n$-th data point of client $c$ (which can be a feature matrix or generally a feature tensor), $N_c$ is the total number of data points at client $c$, $J(\pmb{X}_{c, n}, \pmb{\theta})$ is the loss function used in the FL setting and $f_c(\pmb{\theta})$ is the local loss associated with client $c$ and its data. The overall loss function we optimize is $f(\pmb{\theta})$.

Problem~(\ref{eq1}) is solved using gradient descent. The gradient descent update at iteration $k + 1$ is given by
\begin{equation}
    \pmb{\theta}^{k+1} = \pmb{\theta}^k - \alpha \sum_{c \in \mathfrak{C}} \nabla f_c(\pmb{\theta}^k),
    \label{eq2}
\end{equation}
where $\nabla f_c(\pmb{\theta}^k)$ is the local gradient of client $c$ associated with its data, and $\alpha$ is the learning rate. The sum term in (\ref{eq2}) is a distributed version of gradient descent, also known as \textit{Federated Averaging} \cite{mcmahan2017communication}. Equation~(\ref{eq2}) implies that each client communicates its local gradient to the server at each training iteration. Depending on the quality of the network connection of each client, a significant overhead is introduced to the FL process. This overhead can surpass the computational cost of training a model for the client.
To minimize the data transmission overhead on the distributed training process, we propose to compress the gradient of the loss function, which is reshaped to a matrix or tensor, into a more compact form utilizing a low-rank approximation~\cite{kim2019efficient} \cite{liu2024marvel} \cite{liu2022deep} and then quantize the resulting compact form to reduce further the volume of the data to be transmitted at each iteration. The proposed novel scheme leverages the low-rank approximation of neural network gradients and established quantization algorithms.

The outline of the paper is as follows. Section~\ref{sec2} briefly describes the preliminaries, i.e., gradient compression and quantization. Section~\ref{sec3} details the proposed Quantized Rank Reduction  (QRR) scheme and discusses the experimental results. Conclusions are drawn in Section~\ref{sec4}. The code for QRR can be found at \cite{IARIA2025_QRR_code}.

\section{PRELIMINARIES}\label{sec2}

\subsection{Gradient Compression}

Neural network gradients are expressed in matrix or vector form \cite{clark2017computing}. Suppose we have a function $\pmb{f}: \mathbb{R}^n \rightarrow \mathbb{R}^m$ that maps a vector of length $n$ to a vector of length $m$:
\begin{equation}
    \pmb{f}(\pmb{x}) = \begin{bmatrix}
    f_1(x_1, \ldots, x_n) \\
    f_2(x_1, \ldots, x_n) \\
    \vdots \\
    f_m(x_1, \ldots, x_n)
    \end{bmatrix}.
\end{equation}
The partial derivatives of the vector function are stored in the Jacobian matrix $\frac{\partial\pmb{f}}{\partial\pmb{x}}$, with $\left( \frac{\partial\pmb{f}}{\partial\pmb{x}} \right)_{ij} = \frac{\partial f_i}{\partial x_j}$:
\begin{equation}
    \frac{\partial\pmb{f}}{\partial\pmb{x}} = \begin{bmatrix}
        \frac{\partial f_1}{\partial x_1} & \ldots & \frac{\partial f_1}{\partial x_n} \\
        \vdots & \ddots & \vdots \\
        \frac{\partial f_m}{\partial x_1} & \ldots & \frac{\partial f_m}{\partial x_n}
    \end{bmatrix}.
    \label{eq:Jacobian}
\end{equation}
In the FL context, Jacobian matrices, such as (\ref{eq:Jacobian}), are computed by the clients using the backpropagation algorithm and sent back to the server. The server aggregates them to train the central model via gradient descent. For example, consider the weights of a fully connected layer $\pmb{W} \in \mathbb{R}^{D_{out}\times D_{in}}$ and the bias term $\pmb{b} \in \mathbb{R}^{D_{out} \times 1}$ along with the scalar loss function $J(\cdot)$ used by the neural network, where $D_{out}$ is the size of the fully connected layer output and $D_{in}$ is the size of the input to that layer. After training on its data, each client will derive a gradient reshaped as matrix $\frac{\partial J}{\partial \pmb{W}} \in \mathbb{R}^{D_{out}\times D_{in}}$, as well as the gradient for the bias term $\frac{\partial J}{\partial \pmb{b}} \in \mathbb{R}^{D_{out}\times 1}$. These gradients will be transmitted to the server to train the central model. 

Transmitting the gradients to the server can be slow, especially when training a model with many parameters. The biggest communications overhead comes from $\frac{\partial J}{\partial \pmb{W}}$ and not from $\frac{\partial J}{\partial \pmb{b}}$. This is why we seek to compress $\frac{\partial J}{\partial \pmb{W}}$ by applying the truncated Singular Value Decomposition (SVD), transmitting only the SVD components to the server, and reconstructing $\frac{\partial J}{\partial \pmb{W}}$ on the server using the SVD components.

SVD is a matrix factorization technique that decomposes a matrix $\pmb{A} \in \mathbb{R}^{m\times n}$ into three matrices:
\begin{equation}
    \pmb{A} = \pmb{U} \; \pmb{\Sigma} \; \pmb{V}^\top,
\end{equation}
where $\pmb{U}$ is an $m \times m$ orthonormal matrix containing the left singular vectors of $\pmb{A}$ in its columns, $\pmb{\Sigma}$ is an $m \times n$ matrix with the singular values $\sigma_1, \sigma_2, \ldots, \sigma_r$, in descending order as its diagonal entries, for $r \leq \min(m, n)$ being the rank of matrix $\pmb{A}$, and $\pmb{V}$ is a $n \times n$ orthogonal matrix containing the right singular vectors of $\pmb{A}$ in its columns. We can approximate the matrix $\pmb{A}$ by keeping only the $\nu$ largest singular values:
\begin{equation}
    \pmb{A} \approx \pmb{A}_\nu = \pmb{U}_\nu \; \pmb{\Sigma}_\nu \; \pmb{V}_\nu^\top,
\end{equation}
where $\pmb{U}_\nu \in \mathbb{R}^{m \times \nu}$, $\pmb{\Sigma}_\nu \in \mathbb{R}^{\nu \times \nu}$ and $\pmb{V}_\nu \in \mathbb{R}^{n \times \nu}$ with $\nu < r$. The approximation error of $\pmb{A}$ by $\pmb{A}_\nu$ is given by 
\begin{equation}
    ||\pmb{A} - \pmb{A}_\nu||^2_F = \sum_{j=\nu+1}^r \sigma_j^2,
\end{equation}
where $||\cdot||_F$ denotes the Frobenius norm and $\sigma_j$, $j > \nu$ are the truncated singular values.

\begin{figure}[!t]
    \centering
    \includegraphics[width=0.9\linewidth]{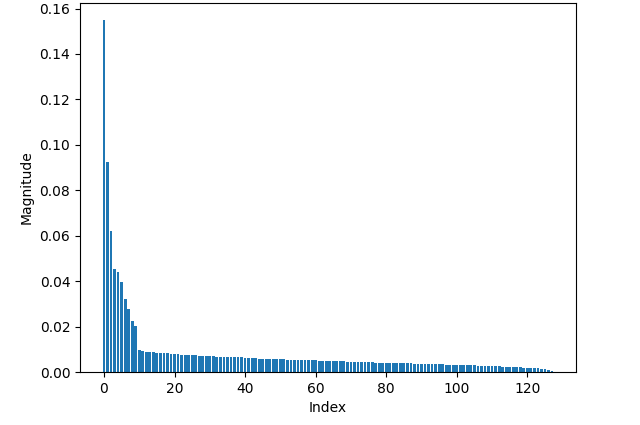}
    \caption{Magnitude of the singular values of the gradient of a fully connected layer.}
    \label{fig:sing_vals}
\end{figure}

The approximation of $\frac{\partial J}{\partial \pmb{W}} \in \mathbb{R}^{D_{out} \times D_{in}}$ with a truncated SVD is justified because such matrices are generally low-rank and have a few dominant singular values \cite{oymak2019generalization}. This was experimentally verified by plotting the magnitudes of the singular values of a fully connected layer's gradient in Figure~\ref{fig:sing_vals}, where only a few of the 128 singular values are significantly larger than 0.

Suppose we only transmit $\pmb{U}_\nu$, $\pmb{V}_\nu$, and the diagonal entries of $\pmb{\Sigma}_\nu$. For the truncated SVD to be more communication-efficient than transmitting the full matrix $\frac{\partial J}{\partial \pmb{W}}$, the following inequality must hold:
\begin{equation}
    \label{eq8}
    D_{out} \cdot \nu + \nu + D_{in} \cdot \nu < D_{out} \cdot D_{in}.
\end{equation}

Factorization can also be applied to convolutional layers. In a convolutional layer, the weights are represented by a 4-dimensional tensor $\euca{W} \in \mathbb{R}^{C_{out} \times C_{in} \times H \times W}$, where $C_{out}$ is the number of output channels, $C_{in}$ is the number of input channels and $H \times W$ is the size of the convolutional filter. The bias terms are represented as a vector $\pmb{b} \in \mathbb{R}^{C_{out} \times 1}$. To reduce the communications overhead of transmitting the gradient of a convolutional layer $\frac{\partial J}{\partial \euca{W}}$ reshaped to a tensor, we factorize the tensor using the Tucker decomposition \cite{tucker1966some}, which has been used for factorization and compression of neural networks \cite{calvi2019compression} \cite{chien2017tensor}.

The Tucker decomposition is a higher-order generalization of SVD. It factorizes a tensor $\euca{X} \in \mathbb{R}^{I_1 \times I_2 \times \ldots \times I_N}$ into a core tensor $\euca{G} \in \mathbb{R}^{r_1 \times r_2 \times \ldots \times r_N}$ and a set of factor matrices $\pmb{F}_i \in \mathbb{R}^{I_i\times r_i}$, $i=1, \ldots, N$, where $r_i < I_i$ are the reduced ranks on each mode.  $\euca{X}$ is approximated as \cite{Lathauwer1997}:
\begin{equation}
     \euca{X} \approx \euca{G} \times_1 \pmb{F}_1 \times_2 \pmb{F}_2 \times_3 \ldots \times_N \pmb{F}_N,
\end{equation}
where $\times_n$ denotes the mode-$n$ product of a tensor and matrix. Given a tensor $\euca{X} \in \mathbb{R}^{I_1 \times I_2 \times \ldots \times I_N}$ and a matrix $\pmb{F} \in \mathbb{R}^{J\times I_n}$ the mode-$n$ product of $\euca{X}$ with $\pmb{F}$ is denoted as $\euca{Y} = \euca{X} \times_n \pmb{F}$, where $\euca{Y} \in \mathbb{R}^{I_1\times \ldots \times I_{n-1} \times J \times I_{n+1} \ldots \times I_N}$ has elements:
\begin{equation}
    \euca{Y}_{i_1, \ldots, i_{n-1}, j, i_{n+1}, \ldots, i_N} = \sum_{i_n=1}^{I_n} \euca{X}_{i_1, \ldots, i_N} \cdot \pmb{F}_{j, i_n}.
\end{equation}

When transmitting the gradient of a convolutional layer reshaped to a tensor,  $\frac{\partial J}{\partial \euca{W}} \in \mathbb{R}^{C_{out} \times C_{in} \times H \times W}$, with reduced ranks for each mode $r_1$, $r_2$, $r_3$, and $r_4$, we only transmit the core tensor and factor matrices. For the Tucker decomposition to be more communication-efficient, the following inequality must be true:
\begin{align}
    \label{eq11}
    & r_1\cdot r_2\cdot r_3\cdot r_4 + C_{out}\cdot r_1 + C_{in}\cdot r_2 \nonumber \\
    &\quad + H\cdot r_3 + W\cdot r_4 < C_{out}\cdot C_{in}\cdot H\cdot W.
\end{align}

\subsection{Gradient Quantization}
\label{sec:quant}

To further reduce the communication overhead of the FL setup, in addition to compressing the updates sent by the clients to the server, we also quantize them. Quantizing the gradients of each client leads to a modified version of (\ref{eq2}) called  Quantized Gradient Descent~\cite{alistarh2017qsgd}:
\begin{equation}
    \pmb{\theta}^{k+1} = \pmb{\theta}^k - \alpha \sum_{c \in \mathfrak{C}} Q \left(\nabla f_c(\pmb{\theta}^k)\right),
\end{equation}
where $Q \left(\nabla f_c(\pmb{\theta}^k)\right)$ is the quantized gradient update of client $c$. Methods employing differential quantization of the gradients have also been proposed \cite{mishchenko2023distributedlearningcompressedgradient} \cite{tonellotto2021neural}. 

The quantization scheme we use resorts to the Lazily Aggregated Quantized (LAQ) algorithm \cite{sun2020lazily}. Specifically, in LAQ, the gradient descent update is given by
\begin{equation}
    \pmb{\theta}^{k+1} = \pmb{\theta}^k - \alpha \:\nabla^k, \; \text{with} \; \nabla^k = \nabla^{k-1} + \sum_{c \in \mathfrak{C}} \delta Q^k_c,
\end{equation}
where $\nabla^k$ is the aggregated quantized gradient updates at iteration $k$, and $\delta Q^k_c := Q_c(\pmb{\theta}^k) - Q_c(\pmb{\theta}^{k-1})$ is the difference of the quantized gradient updates of client $c$ at iterations $k$ and $k - 1$. The quantized gradient update of client $c$ at iteration $k$ is $Q_c(\pmb{\theta}^k)$, and it is computed using the current gradient update $\nabla f_c(\pmb{\theta}^k)$ and the previous quantized update $Q_c(\pmb{\theta}^{k-1})$:
\begin{equation}
    Q_c(\pmb{\theta}^k) = \mathbb{Q}\left(\nabla f_c(\pmb{\theta}^k),\: Q_c(\pmb{\theta}^{k-1})\right),
\end{equation}
where $\mathbb{Q}$ denotes the quantization operator. The operator $\mathbb{Q}$ entails the following quantization scheme. 

The gradient update $\nabla f_c(\pmb{\theta}^k)$ is quantized by projecting each element on an evenly-spaced grid. This grid is centered at $Q_c(\pmb{\theta}^{k-1})$ and has a radius of $R^k_c = ||\nabla f_c(\pmb{\theta}^k) - Q_c(\pmb{\theta}^{k-1})||_\infty$, where $||\pmb{x}||_\infty = \max (|x_1|, \ldots, |x_n|)$ is the max norm. The $i$-th element of the quantized gradient update of client $c$ at iteration $k$ is mapped to an integer as follows
\begin{equation}
    [q_c(\pmb{\theta}^k)]_i = \left\lfloor \frac {[\nabla f_c(\pmb{\theta}^k)]_i - [Q_c(\pmb{\theta}^{k-1})]_i + R^k_c} {2 \tau R^k_c} + \frac{1}{2} \right\rfloor,
\end{equation}
with $\tau := 1 / (2^\beta - 1)$ defining the discretization interval. All $[q_c(\pmb{\theta}^k)]_i$ are integers in the range $\{0, 1, \ldots, 2^\beta - 1\}$ and therefore can be encoded by using only $\beta$ bits. The difference $\delta Q^k_c$ is computed as
\begin{equation}
    \delta Q^k_c = Q_c(\pmb{\theta}^k) - Q_c(\pmb{\theta}^{k-1}) = 2\tau R^k_c \; Q_c(\pmb{\theta}^k) - R^k_c \pmb{1},
\end{equation}
where $\pmb{1} = [1 \ldots 1]^\top$. This quantity can be transmitted with 32 + $\beta \, n$ bits instead of 32$n$ bits. That is, 32 bits for $R^k_c$ and $\beta$ bits for each of the $n$ elements of the gradient update. The computation requires each client to retain a local copy of $Q_c(\pmb{\theta}^{k-1})$. The server can recover the gradient update of client $c$, assuming it knows the number of bits used for quantization,  $\beta$, as
\begin{equation}
    \label{eq17}
    Q_c(\pmb{\theta}^k) = Q_c(\pmb{\theta}^{k-1}) + \delta Q_c^k.
\end{equation}
The discretization interval is $2\tau \: R^k_c$. Therefore, the quantization error cannot be larger than half of the interval
\begin{equation}
    ||\nabla f_c(\pmb{\theta}^k) - Q_c(\pmb{\theta}^k)||_\infty \leq \tau \: R^k_c.
\end{equation}

\section{PROPOSED SCHEME}\label{sec3}

\subsection{Quantized Rank Reduction}

By combining compression and quantization, we propose a new scheme for communication-efficient FL, namely the QRR. The gradient descent step (\ref{eq2}) becomes

\begin{equation}
\resizebox{.9\hsize}{!}{$
    \begin{aligned}
        \pmb{\theta}^{k+1} &= \pmb{\theta}^k - \alpha \sum_{c \in \mathfrak{C}} QRR_c \left(\pmb{\theta}^k\right), \\
        QRR _c\left(\pmb{\theta}^k\right) &= \mathbb{C}^{-1} \left( \mathbb{Q} \left( \mathbb{C} \left( \nabla f_c(\pmb{\theta}^k) \right), \mathbb{C} \left( \nabla f_c(\pmb{\theta}^{k-1}) \right) \right) \right),
    \end{aligned}$}
\end{equation}
where $\mathbb{Q}$ is the quantization operator, $\mathbb{C}$ is the compression operator, and $\mathbb{C}^{-1}$ is the decompression operator. Each client applies the operators $\mathbb{C}$ and $\mathbb{Q}$ to compress and quantize its gradient update, while the server receives the updates and applies $\mathbb{C}^{-1}$ to decompress them and perform gradient descent.

$\mathbb{C}$ entails compressing the gradients using SVD or Tucker decomposition. For the gradient of a fully connected layer of client $c$ at iteration $k$ reshaped to a matrix $\frac{\partial J}{\partial \pmb{W}_c^k} \in \mathbb{R}^{D_{out} \times D_{in}}$ we use a truncated SVD for compression
\begin{equation}
    \frac{\partial J}{\partial \pmb{W}_c^k} \approx \pmb{U}_c^k \; \pmb{\Sigma}_c^k \; (\pmb{V}_c^k)^\top,
\end{equation}
where $\pmb{U}_c^k$, $\pmb{\Sigma}_c^k$ and $\pmb{V}_c^k$ are the SVD components of $\frac{\partial J}{\partial \pmb{W}_c^k}$ retaining only the $\nu$ largest singular values.

In case the gradient update is a tensor, such as the gradient of a convolutional layer $\frac{\partial J}{\partial \euca{W}_c^k} \in \mathbb{R}^{C_{out} \times C_{in} \times H \times W}$,  we compress it using the Tucker decomposition
\begin{equation}
    \frac{\partial J}{\partial \euca{W}_c^k} \approx \euca{G}_c^k \times_1 (\pmb{F}_1)_c^k \times_2 (\pmb{F}_2)_c^k \times_3 (\pmb{F}_3)_c^k \times_4 (\pmb{F}_4)_c^k.
\end{equation}

The compression is controlled by the parameter $p$, which represents the percentage of the original rank that is retained. For SVD, the new reduced rank is computed as
\begin{equation}
    \nu = \left\lceil p \cdot \min(D_{out}, D_{in}) \right\rceil.
\end{equation}
In the case of the Tucker decomposition, the reduced ranks of the core tensor are computed as
\begin{equation}
    \begin{aligned}
        r_1 &= \left\lceil p \cdot C_{out} \right\rceil, &\quad r_2 &= \left\lceil p \cdot C_{in} \right\rceil, \\
        r_3 &= \left\lceil p \cdot H \right\rceil, &\quad r_4 &= \left\lceil p \cdot W \right\rceil.
    \end{aligned}
\end{equation}
For inequalities (\ref{eq8}) and (\ref{eq11}) to hold, we typically want $p$ to be small, i.e., $p< 0.5$.

The gradients of the bias terms $\frac{\partial J}{\partial \pmb{b}_c^k} \in \mathbb{R}^{D_{out} \times 1}$ are quantized only without compression. 

The operator $\mathbb{Q}$ is described in Section \ref{sec:quant}. Each component resulting from the factorization of the gradient update using either SVD or Tucker decomposition is quantized according to this scheme. Client $c$ must store the previous quantized components of its gradient update locally. For each matrix $\frac{\partial J}{\partial \pmb{W}_c^k}$ it has to store $Q(\pmb{U}_c^{k-1})$, $Q(\pmb{\Sigma}_c^{k-1})$ and $Q(\pmb{V}_c^{k-1})$. For each gradient tensor $\frac{\partial J}{\partial \euca{W}_c^k}$, it has to store $Q(\euca{G}_c^{k-1})$ and $Q( (\pmb{F}_1)_c^{k-1})$, $\ldots, Q((\pmb{F}_4)_c^{k-1})$. For each bias  gradient vector $\frac{\partial J}{\partial \pmb{b}_c^k}$ the previous quantized vector $Q(\frac{\partial J}{\partial \pmb{b}_c^{k-1}})$ must also be stored. The parameter $\beta$ is the number of bits used to encode each element and controls the quantization accuracy.

The server receives each client's gradient updates and computes the current iteration's quantized factor components according to (\ref{eq17}). Equation~(\ref{eq17}) requires that the server also store each client's previously quantized factors. Once the server has the current quantized factors, it applies the operator $\mathbb{C}^{-1}$ to reconstruct the gradient updates of each client. That is, for each client $c$ and each model parameter $P$ in the clients' gradient updates,
\begin{flushleft}
    \begin{itemize}
        \item if $P = \pmb{W}_c^k \in \mathbb{R}^{D_{out} \times D_{in}}:$ \\
        \begin{equation}
            \frac{\partial J}{\partial \pmb{W}_c^k} \approx Q(\pmb{U}_c^k) \; Q(\pmb{\Sigma}_c^k) \: Q(\pmb{V}_c^k)^\top,
        \end{equation}
    
        \item if $P = \euca{W}_c^k \in \mathbb{R}^{C_{out} \times C_{in} \times H \times W}:$ \\
        \begin{equation}
            \begin{aligned}
                \frac{\partial J}{\partial \euca{W}_c^k} \approx Q(\euca{G}_c^k) &\times_1 Q((\pmb{F}_1)_c^k) \times_2 
                Q((\pmb{F}_2)_c^k) \\ &\times_3  Q((\pmb{F}_3)_c^k) \times_4  Q((\pmb{F}_4)_c^k),
            \end{aligned}
        \end{equation}
    
        \item if $P = \pmb{b}_c^k \in \mathbb{R}^{D_{out} \times 1}:$ \\
        \begin{equation}
            \frac{\partial J}{\partial \pmb{b}_c^k} \approx Q(\frac{\partial J}{\partial \pmb{b}_c^k}).
        \end{equation}
    \end{itemize}
\end{flushleft}
The server then uses the gradient approximations to perform the distributed gradient descent.

\subsection{Experimental Results}
Experiments were conducted to compare the performance of the proposed QRR with stochastic federated averaging, referred to as Stochastic Gradient Descent (SGD), and with the Stochastic LAQ (SLAQ)\cite{sun2020lazily}. To measure the performance of each method, we kept track of the loss and accuracy of the model, as well as the number of bits transmitted by the clients during each iteration. Since the SLAQ algorithm skips uploading the gradient update of some clients based on their magnitude, we also recorded the number of communications. Next, we clarify the terms used in the experiments:
\begin{itemize}
    \item By \textbf{iteration}, we mean a full round of FL, which consists of the server passing the central model’s weights to the clients, the clients computing their local mean gradient over a single batch and sending it to the server, and the server aggregating the clients’ gradients and updating the central model.

    \item By \textbf{communication}, we refer to the data exchange from the client to the server,  i.e., when the client sends its local gradient update to the server.

    \item By \textbf{bits}, we measure only the number of bits of the gradient updates transferred from the clients to the server, since the bits required to transmit the model weights from the server to all the clients are constant and common across all methods.
\end{itemize}

All the experiments used 10 clients and quantized the compressed gradient updates using $\beta = 8$ bits. The learning rate was $\alpha = 0.001$, and the batch size was equal to 512. For the SLAQ algorithm, the parameters used were $D = 10$, $\xi_1, \ldots, \xi_D = 1 / D$, and 8 bits for quantization.

The first experiment utilized the MNIST dataset \cite{deng2012mnist} of 28~$\times$~28 grayscale images of handwritten digits. A simple Multi-Layer Perceptron (MLP) network was employed, comprising a hidden layer with 200 neurons, a Rectified Linear Unit (ReLU) activation function, and input and output layers of size 784 ($28 \times 28$) and 10, respectively, with a cross-entropy loss function. 60,000 training samples were randomly selected and equally distributed among the 10 clients. A total of 10,000 test samples were used to evaluate the performance of the central model. The results for 1000 iterations are presented in Table~\ref{tab:mlp_res} for various values of $p$ in QRR.

\begin{table*}[!ht]
    \centering
    \caption{RESULTS OF QRR COMPARED TO SLAQ and SGD FOR AN MLP APPLIED TO THE MNIST DATASET.}
     \resizebox{1.4\columnwidth}{!}{%
    \begin{tabular}{|c|c|c|c|c|c|c|}
        \cline{1-7}
         \textbf{Algorithm} & \textbf{\# Iterations} & \textbf{\# Bits} & \textbf{\# Communications} & \textbf{Loss} & \textbf{Accuracy} & \textbf{Gradient $\ell_2$ norm} \\ \hline
         SGD & 1000 & $5.088\times10^{10}$ & 10000 & 0.376 & 89.92\% & 2.297 \\ \hline
         SLAQ & 1000 & $1.089\times10^{10}$ & 8559 & 0.378 & 89.89\% & 2.026 \\ \hline
         QRR($p=0.3$) & 1000 & $4.798\times10^9$ & 10000 & 0.415 & 89.20\% & 1.945 \\ \hline
         QRR($p=0.2$) & 1000 & $3.205\times10^9$ & 10000 & 0.441 & 88.93\% & 2.846 \\ \hline
         QRR($p=0.1$) & 1000 & $1.612\times10^9$ & 10000 & 0.501 & 88.22\% & 1.866 \\ \hline
    \end{tabular}
    }
    \label{tab:mlp_res}
\end{table*}

\begin{figure}[!ht]
    \centering
    \resizebox{0.5\textwidth}{!}{
        \begin{tabular}{cc}
           \subfloat[][Loss vs. iterations]{\includegraphics[width=0.32\textwidth]{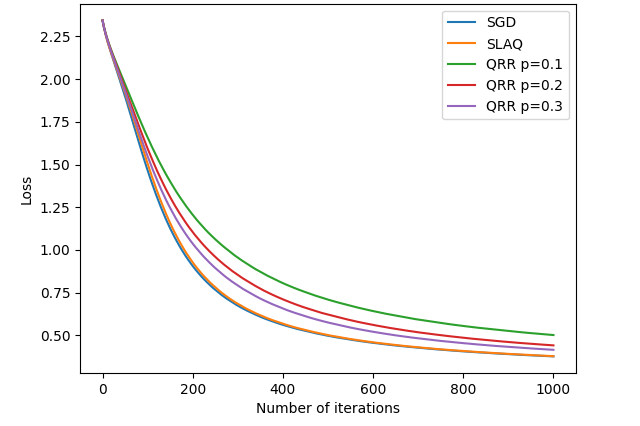} \label{fig:mlp_res:a}} &
            \subfloat[][Loss vs. Bits]{\includegraphics[width=0.32\textwidth]{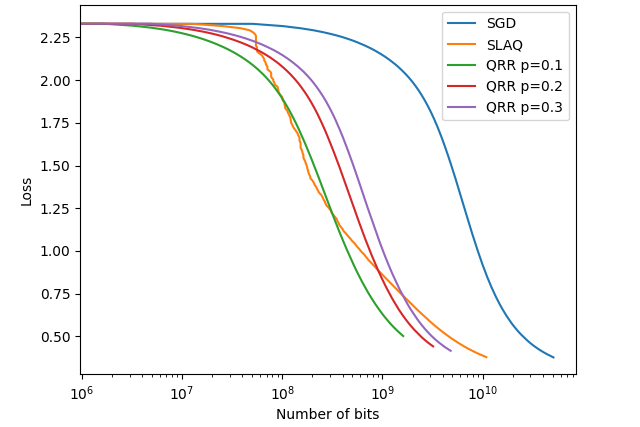} \label{fig:mlp_res:b}} \\
            
            \subfloat[][Gradient $\ell_2$ norm vs. Iterations]{\includegraphics[width=0.32\textwidth]{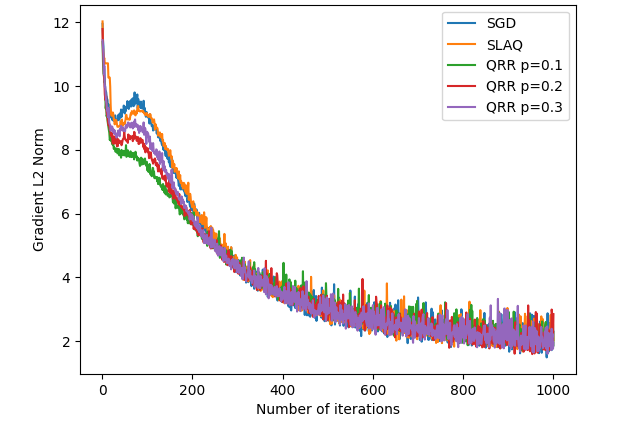} 
            \label{fig:mlp_res:c}} &
            \subfloat[Gradient $\ell_2$ norm vs. Bits]{\includegraphics[width=0.32\textwidth]{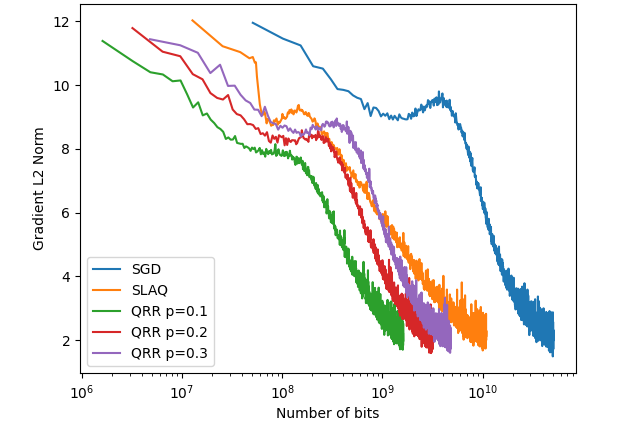}  \label{fig:mlp_res:d}} \\
            
            \subfloat[][Accuracy vs. Iterations]{\includegraphics[width=0.32\textwidth]{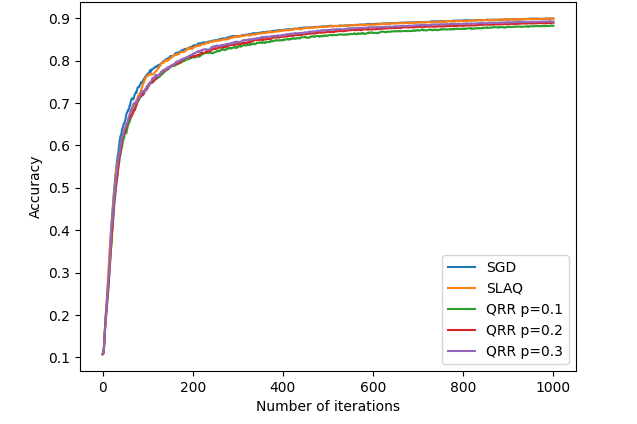}  \label{fig:mlp_res:e}} &
            \subfloat[][Accuracy vs.  Bits]{\includegraphics[width=0.32\textwidth]{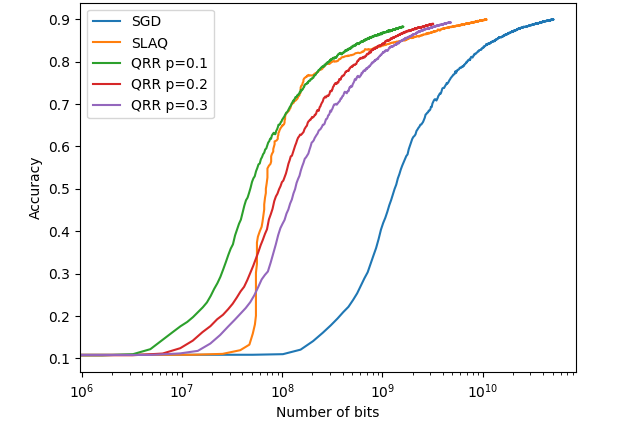}  \label{fig:mlp_res:f}}
        \end{tabular}
    }
    \caption{Loss, gradient $\ell_2$ norm, and accuracy plotted against the number of iterations and bits for the MLP network and the MNIST dataset.}
    \label{fig:mlp_res}
\end{figure}

QRR achieves an accuracy of around 1-2\% lower than SGD and SLAQ. However, it transmits 3.16-9.43\% of the bits transmitted by SGD and 14.8-44.05\% of the bits transmitted by SLAQ, depending on the choice of the parameter $p$. In Figure~\ref{fig:mlp_res}, the loss, the gradient $\ell_2$ norm, and the accuracy are plotted against each method's number of iterations and bits. QRR has a slower convergence rate with respect to (wrt) the iteration number than SGD and SLAQ. The smaller $p$ is, the slower the loss convergence, as evidenced in Figure~\ref{fig:mlp_res:a} since we have less accurate reconstructions of the gradients with smaller $p$ values. However, performance wrt the number of bits transmitted is better, as seen in Figures~\ref{fig:mlp_res:b}, \ref{fig:mlp_res:d}, and~\ref{fig:mlp_res:f}, i.e., a smaller loss, a smaller gradient $\ell_2$ norm, and higher accuracy are measured for a fixed number of bits.

The second experiment used the same setup as the first, with the difference that the MLP network was replaced by a Convolutional Neural Network (CNN). The CNN consisted of 2 convolutional layers using 3$\times$3 filters with 16 and 32 output channels, respectively, a max pooling layer that reduced the input size by half, and 1 fully connected layer. The activation function used after each layer was the ReLU function, and the loss function used was the cross-entropy loss.

\begin{figure}[!ht]
    \centering
    \resizebox{0.5\textwidth}{!}{
        \begin{tabular}{cc}
            \subfloat[Loss vs. Iterations]{\includegraphics[width=0.32\textwidth]{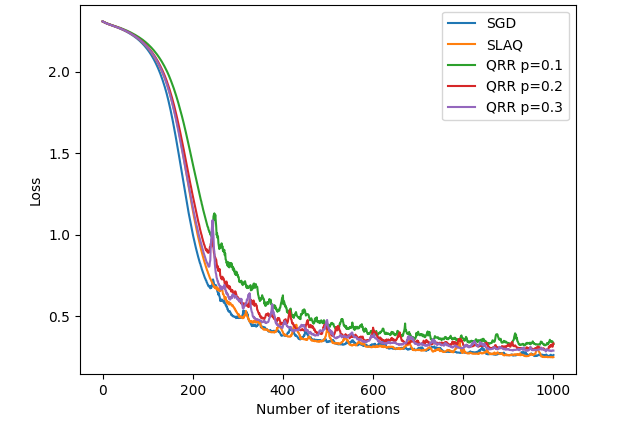}} &
            \subfloat[Loss vs. Bits]{\includegraphics[width=0.32\textwidth]{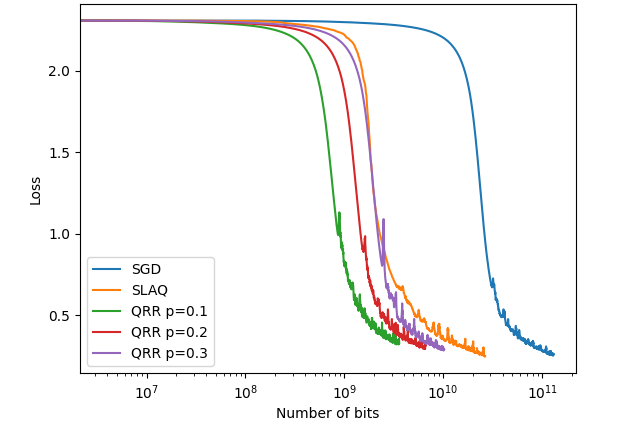}} \\
            
            \subfloat[Gradient $\ell_2$ norm vs. Iterations]{\includegraphics[width=0.32\textwidth]{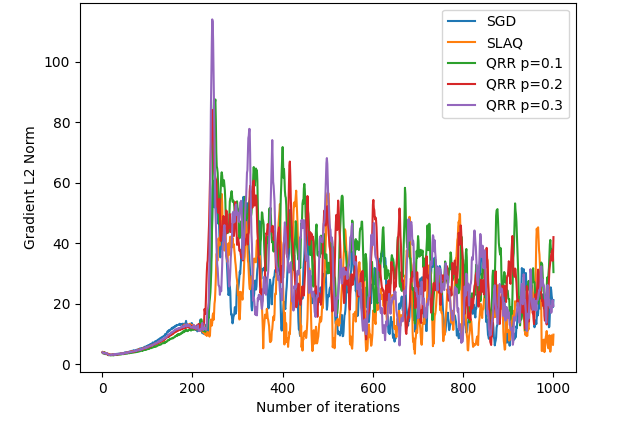}} &
            \subfloat[Gradient  $\ell_2$ norm vs. Bits]{\includegraphics[width=0.32\textwidth]{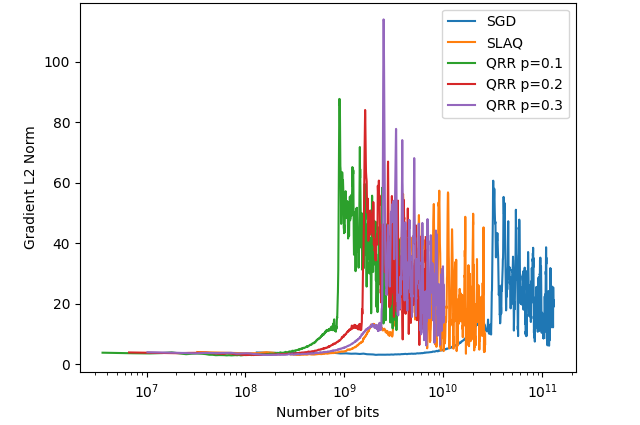}} \\
            
            \subfloat[Accuracy vs. Iterations]{\includegraphics[width=0.32\textwidth]{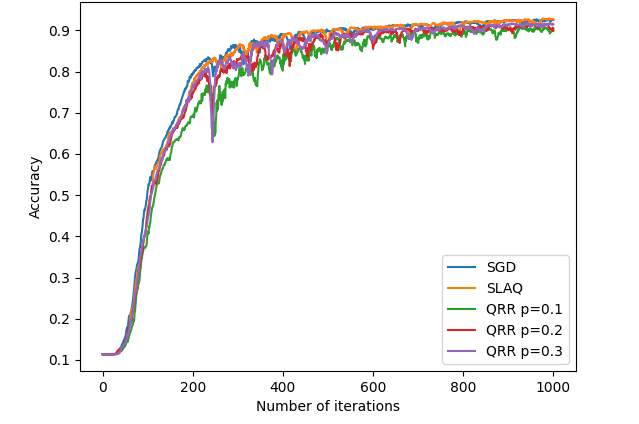}} &
            \subfloat[Accuracy vs. Bits]{\includegraphics[width=0.32\textwidth]{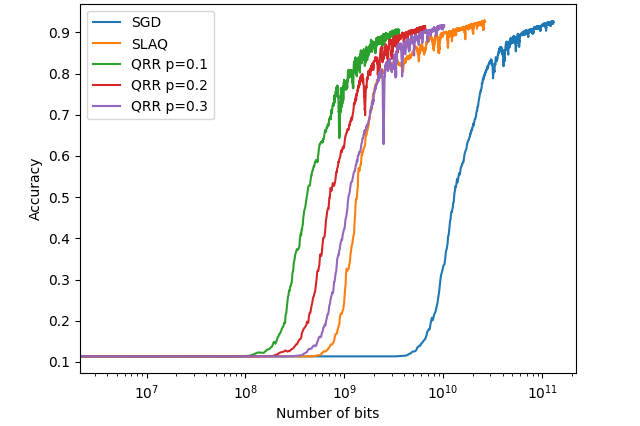}}
        \end{tabular}
    }
    \caption{Loss, gradient $\ell_2$ norm, and accuracy plotted against the number of iterations and bits for the CNN and the MNIST dataset.}
    \label{fig:cnn_res}
\end{figure}

Table~\ref{tab:cnn_res} summarizes the results using the CNN. Figure~\ref{fig:cnn_res} displays the evolution of the loss, gradient $\ell_2$ norm, and accuracy wrt the number of iterations and bits. The curves for loss and accuracy versus iterations or bits are similar to those of the first experiment. QRR scores 1-3\% lower in accuracy but requires 2.75-7.84\% of the bits of SGD and 13.52-38.52\% of the bits of SLAQ, depending on the choice of $p$.

\begin{table*}[!ht]
    \centering
    \caption{RESULTS OF QRR COMPARED TO SLAQ and SGD FOR A CNN APPLIED TO THE MNIST DATASET.}
     \resizebox{1.4\columnwidth}{!}{%
    \begin{tabular}{|c|c|c|c|c|c|c|}
        \cline{1-7}
         \textbf{Algorithm} & \textbf{\# Iterations} & \textbf{\# Bits} & \textbf{\# Communications} & \textbf{Loss} & \textbf{Accuracy} & \textbf{Gradient $\ell_2$ norm} \\ \hline
         SGD & 1000 & $1.302\times10^{11}$ & 10000 & 0.263 & 92.56\% & 21.154 \\ \hline
         SLAQ & 1000 & $2.653\times10^{10}$ & 8151 & 0.251 & 92.70\% & 9.769 \\ \hline
         QRR($p=0.3$) & 1000 & $1.022\times10^{10}$ & 10000 & 0.291 & 91.49\% & 19.287 \\ \hline
         QRR($p=0.2$) & 1000 & $6.650\times10^9$ & 10000 & 0.335 & 89.91\% & 42.026 \\ \hline
          QRR($p=0.1$) & 1000 & $3.588\times10^9$ & 10000 & 0.330 & 90.48\% & 30.455 \\ \hline
    \end{tabular}
    }
    \label{tab:cnn_res}
\end{table*}

In the third experiment, the CIFAR-10 dataset \cite{Krizhevsky2009LearningML} was used with a small, VGG-like \cite{simonyan2015deepconvolutionalnetworkslargescale} CNN consisting of three convolutional blocks with 3×3 convolutions, ReLU activations, max pooling, and dropout layers, with the number of filters increasing from 32 to 64 and then to 128. We used different values of $p$ to demonstrate that $p$ can be chosen based on the client's connection speed and the amount of data transmitted from that client. Evenly spaced values in $[0.1, 0.3]$ were assigned to the $p$ parameter of each client. The experiment ran for 2000 iterations, using a learning rate of 0.01 for the first 1000 iterations to accelerate convergence, and then 0.001 for the remaining iterations to ensure stable training.

\begin{figure}[!ht]
    \centering
    \resizebox{0.5\textwidth}{!}{
        \begin{tabular}{cc}
            \subfloat[Loss vs. Iteration]{\includegraphics[width=0.32\textwidth]{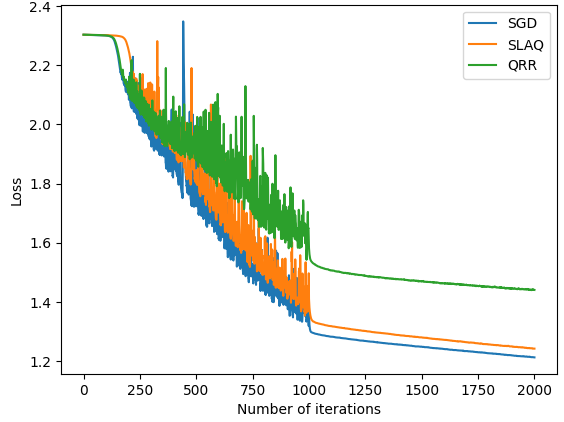}} &
            \subfloat[Loss vs. Bits]{\includegraphics[width=0.32\textwidth]{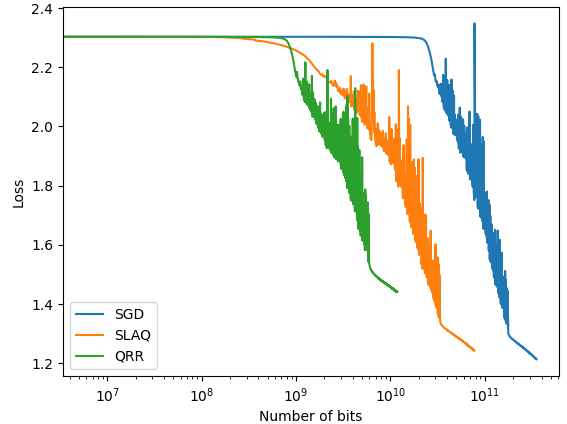}} \\
            
            \subfloat[Gradient vs. Iteration]{\includegraphics[width=0.32\textwidth]{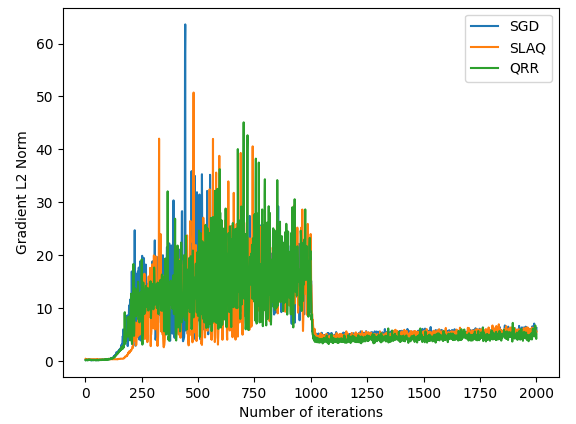}} &
            \subfloat[Gradient vs. Bits]{\includegraphics[width=0.32\textwidth]{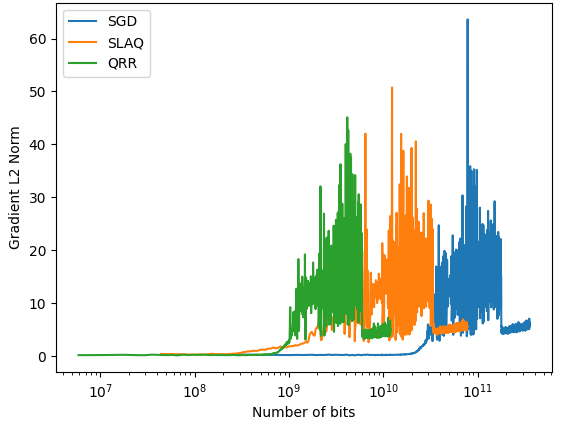}} \\
            
            \subfloat[Accuracy vs. Iteration]{\includegraphics[width=0.32\textwidth]{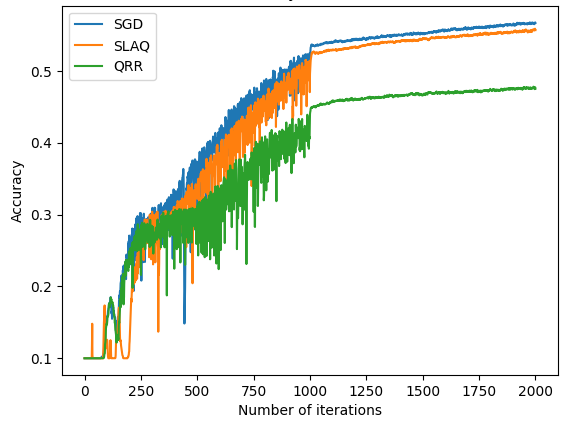}} &
            \subfloat[Accuracy vs. Bits]{\includegraphics[width=0.32\textwidth]{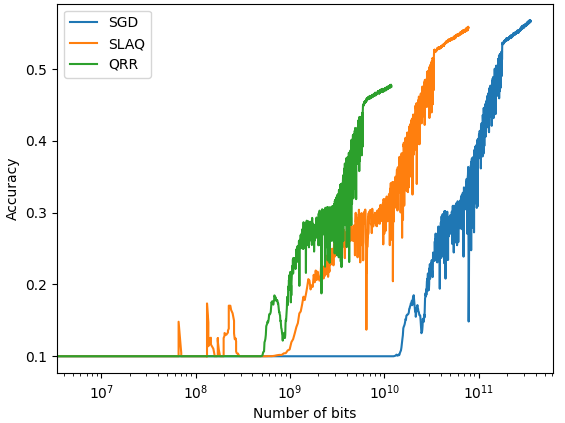}}
        \end{tabular}
    }
    \caption{Loss, gradient $\ell_2$ norm, and accuracy plotted against the number of iterations and bits for the VGG-like CNN and the CIFAR-10 dataset.}
    \label{fig:cifar10_res}
\end{figure}

Table~\ref{tab:cifar10_res} shows that QRR achieves 8–9\% lower accuracy than SGD and SLAQ, while transmitting only 3.34\% and 15.26\% of the bits transmitted by SGD and SLAQ, respectively. Figure~\ref{fig:cifar10_res} plots the loss, gradient $\ell_2$ norm, and accuracy versus iterations or transmitted bits for the VGG-like CNN on CIFAR-10. Although the low-rank approximation of the gradients leads to reduced accuracy on this dataset, which is more complex than MNIST, QRR remains useful for quickly reaching a deployable model state before switching to a more accurate one, compared to less network-efficient methods such as SGD or SLAQ.

Finally, the client-side overhead of QRR was measured in the setup of the last experiment using SGD as a baseline. On average, QRR needed 1.2$\times$ more memory and 3.82$\times$ more computation time. For comparison, SLAQ required 13$\times$ more memory and 1.08$\times$ more computation time.

\begin{table*}[!ht]
    \centering
    \caption{RESULTS OF QRR COMPARED TO SLAQ and SGD FOR A VGG-LIKE CNN APPLIED TO THE CIFAR-10 DATASET.}
    \resizebox{1.4\columnwidth}{!}{%
    \begin{tabular}{|c|c|c|c|c|c|c|}
        \cline{1-7}
         \textbf{Algorithm} & \textbf{\# Iterations} & \textbf{\# Bits} & \textbf{\# Communications} & \textbf{Loss} & \textbf{Accuracy} & \textbf{Gradient $\ell_2$ norm} \\ \hline
         SGD & 2000 & $3.52\times10^{11}$ & 20000 & 1.213 & 56.72\% & 6.246 \\ \hline
         SLAQ & 2000 & $7.72\times10^{10}$ & 17548 & 1.242 & 55.73\% & 5.493 \\ \hline
         QRR & 2000 & $1.17\times10^{10}$ & 20000 & 1.441 & 47.57\% & 5.088 \\ \hline
    \end{tabular}
    }
    \label{tab:cifar10_res}
\end{table*}

\section{Conclusions}\label{sec4}

We have proposed a scheme that leverages the low-rank approximation of neural network gradients and utilizes established quantization algorithms to significantly reduce the amount of data transmitted in an FL setting. The proposed Quantized Rank Reduction scheme has slightly lower accuracy than Federated Averaging or SLAQ, but it transmits only a fraction of the bits required by the other methods. It converges more slowly with the number of iterations, but faster when considering the number of bits transmitted. There is an added computational and memory overhead on both the client and server sides. However, this scheme can prove helpful in network-critical applications, where sensors or devices participating in the distributed learning process are located in remote locations with very slow network connections.

\bibliographystyle{IEEEtran}
\bibliography{References}

\end{document}